\documentclass{article}

\usepackage{microtype}
\usepackage{graphicx}
\usepackage{subfigure}
\usepackage{booktabs} 
\usepackage{enumitem}
\usepackage{hyperref}
\usepackage{dblfloatfix}
\usepackage{color}
\definecolor{darkgreen}{rgb}{0.0, 0.5, 0.0}

\usepackage[accepted]{icml2024}

\usepackage{amsmath}
\usepackage{amssymb}
\usepackage{mathtools}
\usepackage{amsthm}

\usepackage[capitalize,noabbrev]{cleveref}

\theoremstyle{plain}

\theoremstyle{definition}

\theoremstyle{remark}

\usepackage{cuted}

\usepackage[textsize=tiny]{todonotes}

\icmltitlerunning{Submission and Formatting Instructions for GRaM Workshop at ICML 2024}

\begin{document}

\twocolumn[
\icmltitle{The NGT200 Dataset\\ Geometric Multi-View Isolated Sign Recognition}


\icmlsetsymbol{equal}{*}

\begin{icmlauthorlist}
\icmlauthor{Oline Ranum}{yyy}
\icmlauthor{David Wessels}{xxx}
\icmlauthor{Gomèr Otterspeer}{yyy}
\icmlauthor{Erik J Bekkers}{xxx}
\icmlauthor{Floris Roelofsen}{yyy}
\icmlauthor{Jari I. Andersen}{yyy}
\end{icmlauthorlist}

\icmlaffiliation{yyy}{SignLab Amsterdam UvA}
\icmlaffiliation{xxx}{AMLab UvA}

\icmlcorrespondingauthor{Oline Ranum}{o.a.ranum@uva.nl}

\icmlkeywords{Machine Learning, ICML}

\vskip 0.3in
\editorsListText
\vskip 0.3in
]



\printAffiliationsAndNotice{\icmlEqualContribution} 






\begin{abstract}
Sign Language Processing (SLP) provides a foundation for a more inclusive future in language technology; however, the field faces several significant challenges that must be addressed to achieve practical, real-world applications. This work addresses multi-view isolated sign recognition (MV-ISR), and highlights the essential role of 3D awareness and geometry in SLP systems.  We introduce the NGT200 dataset, a novel spatio-temporal multi-view benchmark, establishing MV-ISR as distinct from single-view ISR (SV-ISR). We demonstrate the benefits of synthetic data and propose conditioning sign representations on spatial symmetries inherent in sign language. Leveraging an SE(2) equivariant model improves MV-ISR performance by \texttt{8\%}-\texttt{22\%}  over the baseline.

\end{abstract}

\section{Introduction}
Sign languages (SL) are dynamic, visual and natural languages articulated using the hands, face, and body. They are expressed through the synthesis of three-dimensional shapes, structures and movements, and leverage temporal and geometric positioning to convey meaning.

Over the past years, the automatic understanding, processing, and analysis of sign languages have gathered an accelerating amount of attention \cite{koller2020, Rastgoo_2021}. Consequentially, SLP has emerged as a diverse research area, encompassing expertise from varying fields including computer vision, natural language processing (NLP), computer graphics, linguistics, human-computer interaction, and Deaf culture \cite{bragg2019}.

SLP applications encompass services with automated SL accommodation, such as SL smart assistants and machine translation (SL-MT). However, despite noticeable advances within the field, SLP methods lag behind other NLP technologies \cite{yin2021}. Advancing SLP relies on addressing several key challenges, including the lack of large-scale, high-quality datasets, difficulties in generalizing to new signers and situations, and the need for methods that can handle the structural complexities and visual features of sign languages \cite{Joksimoski2022, desai2024}. Moreover, SL linguistics is a young field, with foundational research starting in the 1960s \cite{Stokoe1960}, leaving much still to be understood.



Sign Language Recognition (SLR) methods interpret SL from videos and are crucial for many SLP applications. However, a gap remains between research advancements and real-world deployment, largely due to the reliance on datasets that capture SL from a single, frontal view.  This two-dimensional representation of a three-dimensional language leads to information loss, making variations in viewing angles significantly impact SLR performance.



In daily interactions, such as group conversations or crowded areas, signing is often perceived from multiple angles, making it essential for SLR systems to process signs from varying viewpoints. Additionally, when processing signing from individuals with significant cognitive or functional impairments, imposing stringent regulations on assistive tool usage can be both insensitive and inconvenient. Both scenarios necessitate user-friendly, view-invariant systems to ensure seamless interactions. Therefore, we argue that viewpoints matter in real-world SLR applications.


We take a step towards multi-view SLP by introducing a publicly available multi-view isolated sign dataset and perspectives on multi-view isolated sign recognition (MV-ISR). MV-ISR/SLR is challenging due to the need for models to generalize across signer appearance, articulation style, and viewpoints, compounded by the scarcity of multi-view data. Consequently, MV-SLR algorithms are increasingly compelled to learn more efficiently from limited datasets. 

Multi-view SLP introduces several critical questions. A primary concern is how viewpoint transformations affect SLP accuracy. If the impact proves significant, it becomes essential for the community to explore ethical, consistent, and scalable approaches for incorporating multi-view representations into SL data, while also enhancing the view-invariance of SL models. However, incorporating multi-view geometry increases the task complexity. This prompts a key question: how can we ground and condition SLR architectures to increase sample efficiency, reduce model complexity, and enhance generalizability in response to the challenges posed by multi-view SLR?

To begin addressing these questions, we introduce the NGT200 dataset detailed in Section \ref{Sec:Datasetintro}, which is designed as a preliminary step to explore the multi-view geometry of signing. NGT200 includes 2D landmarks extracted from video clips of isolated signs depicting both human and synthetic signers captured from multiple views. The dataset aligns with the 3D-LEX dataset \cite{ranum2024}, providing 3D ground truth for each sign in the vocabulary. We also release a subsection of the corresponding video data.

We construct \textit{geometric sign graphs} from NGT200 landmarks. Each node $i$ corresponds to a landmark $x_i$ on the human body, with spatial proximity to nodes $j$ corresponding to landmarks $x_j$. The spatial edges are configured to approximate the human bone structure. Sign graphs are lower-dimensional representations of the dynamic geometric shapes and structures formed by the articulators of signs.

In Section \ref{Sec:mvisr}, we use the sign graphs to characterize MV-ISR by its distinction from SV-ISR, and define the task as the challenge of achieving view-invariant predictions of sign language word labels (glosses) from multi-view isolated sign data. In Section \ref{sec:synthdatasec}, we contribute to scalable multi-view data production by demonstrating the efficacy of including synthetic poses in the training data. In Section \ref{Sec:GeomSLR} we propose a method to address the spatial complexity inherent in SL, which is further improved by including multiple views in the dataset. Our approach leverages geometrically grounded models to generate representations that maintain the symmetries intrinsic to the graphs constructed from SL data, including rotational symmetries corresponding to various perspectives. By integrating these inductive biases, the geometric properties are preserved in the representation, which has been shown to improve performance in downstream tasks \cite{wessels2024}.


Models that focus on learning geometrically grounded representations have led to state-of-the-art outcomes in diverse areas such as protein structure prediction \cite{Jumper2021, Baek2021}, n-body simulations \cite{bekkers2024fast}, and 3D-modeling \cite{heidari2024geometric}.
The geometric nature of sign graphs suggests that GDL tools used for shape and geometry analysis (\textit{e.g.} molecular conformations) could significantly impact SLP. In this work, we demonstrate the potential of using equivariant models to address the complexities caused by variations in articulation style and prosodic factors, such as sign amplitude, thereby enhancing the understanding of local symmetries in neural sign representations. 

\textit{Geometric SLP} presents a novel challenge with NGT200 as a new benchmark for the geometry-grounded machine learning community, characterized by the search for patterns and structures governed by linguistic rules and spatio-temporal dependencies. The contributions of this work can be summarized as follows:

\begin{enumerate}[noitemsep]
    \item We introduce a new dataset and benchmark for the task of MV-ISR: \textit{The NGT200 Dataset}. \vspace{2mm}
    \item We provide a proof-of-concept demonstrating that MV-ISR is a distinct task from SV-ISR, necessitating the adoption of novel and more efficient approaches.\vspace{2mm}
    \item We demonstrate that avatar-based synthetic pose data can be used to upscale low-resource MV datasets.  \vspace{2mm} 
    \item We propose leveraging a geometrically informed model to tackle the MV-ISR task, demonstrating significant improvements in gloss prediction accuracy.
\end{enumerate}


\section{Background}

\subsection{Sign Languages and Visual Linguistics}\label{sl-lin}

Sign languages are visual, natural languages with unique structures, grammars, and lexicons. They primarily function as the main languages within Deaf communities, where they emerge and continuously evolve \cite{Padden_2005, Leigh2022}. Additionally, sign languages are utilized in various forms by hard-of-hearing persons, Children/Siblings of Deaf Adults (CODA/SODA), SL interpreters, second language learners and individuals with cognitive and/or physical disorders that impact (spoken) language learning abilities. Hundreds of sign languages are thought to exist worldwide \cite{Eberhard2022},  though their prevalence, accessibility, and lawful recognition vary from country to country \cite{meulder2015, Murray2020}.

Sign languages convey meaning through a collection of asynchronous visual information cues, expressed with manual (hands, arms, fingers) and non-manual (\textit{e.g.} facial expressions, gaze direction, torso, head posture) articulators \cite{Brentari2018}. The basic independent meaningful unit is generally a sign language word. When considered in isolation, the structure of a sign word can largely be characterized in terms of its phonological features: handshape, place of articulation, movement and palm orientation \cite{Stokoe1960}.

\newpage
In continuous signing scenarios, such as sentence construction or conversations, the linguistic landscape transforms, as new linguistic phenomena are introduced at the suprasegmental level.  While the NGT200 dataset is exclusively comprised of sign language words, its important to acknowledge that continuous sign features render the generalization of methods from isolated to continuous SLP a nonlinear and nontrivial process. Examples of such features include co-articulation, where attributes of a sign are influenced by adjacent signs, and increased variation in articulation speed and amplitude as a means to mark prosody. 




\subsection{Sign Language Recognition}
Sign Language Recognition (SLR) is the task of automatically recognizing and interpreting sign language from videos or other motion capture data. The task is commonly divided into Isolated Sign Recognition (ISR) and Continuous Sign Language Recognition (CSLR). ISR focuses on predicting glosses \cite{Sehyr2021, Athitsos2008, Kezar_2023, Jose_2018, Dongxu2020}, by considering visual features from videos, poses and depth estimates. CSLR is the task of recognizing and interpreting entire sign language sentences from SL-corpora 
\cite{forster-etal-2014, Argis_2010, sCHEMBRI_2013}. For an extensive overview of methods and state-of-the-art in SLR see \citet{koller_surv_2020} or \citet{RASTGOO2021113794}. For an extensive summary of sign language datasets, see \citet{kopf_maria_2021}.

\subsection{Multi-View and 3D-aware Sign Language Recognition}
Despite its practical importance and potential to enhance three-dimensional fidelity in SLP tasks, MV-SLR has received little attention in the literature. While sign languages are inherently three-dimensional, most research has focused on two-dimensional projections like single-view videos. However, an emerging body of literature indicates that 3D-awareness matters in SLP. 

The study by \citet{Watkins2024} demonstrates that viewing angle significantly influences human SL recognition, suggesting that sign features transform substantially under rotation, a factor relevant to machine recognition. Additionally, other studies have found that neural networks are sensitive to the three-dimensional linguistic structures of sign language \cite{MartinezRodriguez2023}, and conditioning neural models on these structures improves recognition accuracy \cite{Kezar_2023}.

\citet{Gao2023} provides a proof-of-concept for the importance of viewing angle in SLR.  They produced a multi-view Chinese Sign Language (CSL) dataset with 14 signers and 50 sign classes. Using a Multi-View Knowledge Transfer (MVKT) model, they showed a recognition accuracy drop of over \texttt{50\%} when trained on frontal views and tested on frontal and side views, respectively. They also showed that training with multiple views consistently improved accuracy.

There is a growing trend in SL data production to include multiple views. The How2Sign dataset \cite{Duarte_2021} offers over 80 hours of American Sign Language videos, including speech, English transcripts, RGB-D videos, key points, with both frontal and side views for each recording. A three-hour subset was recorded in a Panoptic studio, enabling detailed 3D pose estimations. Additionally, the fable1 dataset was recently released, a small-scale corpus comprising continuous SL fairy tales in German Sign Language recorded from 7 
viewpoints \cite{nunnari2024}. Both datasets comprise SL sentences, while the NGT200 dataset consists of SL words.



\section{The NGT200 Dataset}\label{Sec:Datasetintro}
We begin by introducing NGT200, containing pose and video data for 200 common NGT signs, captured from three viewpoints with both human and synthetic signers.

\subsection{Vocabulary Construction and Resource Alignment}\label{Sec:Vocab}
The vocabulary of NGT200 is aligned with the SignBank NGT Lexicon \cite{Crasborn2020NGT} and the 3D-LEX dataset \cite{ranum2024}. SignBank NGT is an extensive database that provides detailed linguistic information for individual signs, including phonetic characteristics such as handshapes and handedness. Furthermore, the lexicon provides an additional frontal-view example video for each sign. The 3D-LEX dataset contains 3D motion capture data for the NGT200 vocabulary, providing a 3D ground truth for the NGT200 dataset and enabling the sampling of synthetic data from novel views using an avatar. A comparison between the NGT200 dataset and other isolated sign datasets is provided in Appendix \ref{sec:comparison}.

\subsection{Data Capture}\label{Sec:capture}
\begin{figure}[!t]
\vskip 0.2in
\begin{center}
\centerline{\includegraphics[width=0.7\columnwidth]{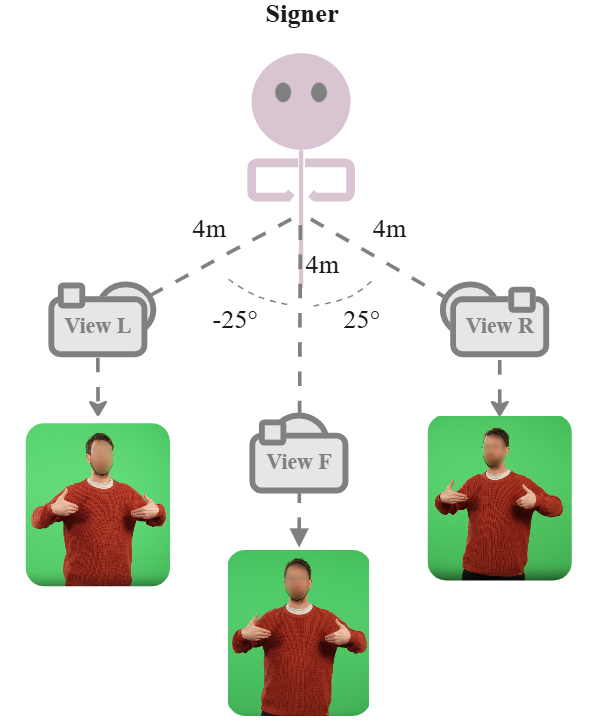}}
\caption{Configuration of video capture setup with the signCollect platform: each camera is positioned 4 meters away from the signer, with a 25$^\circ$ separation between cameras.}
\label{fig:signCollect}
\end{center}
\vskip -0.2in
\end{figure}

The pose data is obtained from a collection of multi-view videos of signers performing sign words. The videos are captured using the signCollect platform, developed by \citet{Otterspeer2024}. 
The signCollect system is a sign recording platform designed to provide a ‘touchless’ interface for sign capture, enabling system operation through gesture recognition. This platform automates the sign collection workflow to efficiently sample signs from multiple viewing angles. 
The sign collection setup for the NGT200 dataset is displayed in Figure \ref{fig:signCollect}, showcasing how the three views are captured from a left, front and right perspective at respectively \texttt{-25$^\circ$}, \texttt{0} and \texttt{25$^\circ$} degrees apart. All three cameras are synchronously triggered to start capturing upon detection of the initialization gesture, ensuring temporal alignment between the different video clips of each view.

\subsection{Spatio-Temporal Point Cloud Construction}\label{Sec:stpc}
\begin{figure}[!t]
\vskip 0.2in
\begin{center}
\centerline{\includegraphics[width=0.78\columnwidth]{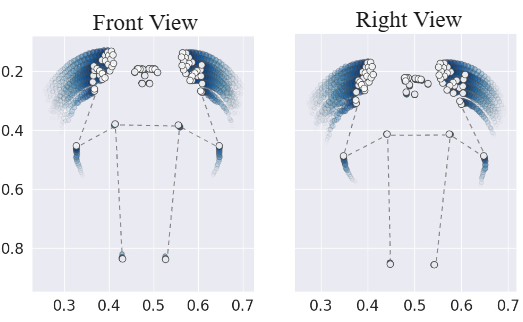}}
\caption{Spatio-temporal point clouds extracted with MediaPipe, displaying the front and right view. White landmarks represent a single frame, while blue landmarks indicate temporal dynamics across multiple frames. Dashed lines connect the landmarks purely for visual enhancement and do not reflect elements in the dataset. }
\label{fig:stpc}
\end{center}
\vskip -0.2in
\end{figure}

We use Holistic MediaPipe version 0.10.11 \cite{mediapipe} to extract landmarks from the videos, as illustrated in Figure \ref{fig:stpc}. This comprehensive framework extracts pose, face, and hand key points, enabling the analysis of full-body gestures, poses, and actions. We extract 11 landmarks from the face, 14 landmarks from the body, and 21 landmarks from each hand per frame. These landmarks are combined such that each sign is represented by a matrix of size $T_c\times N_{lm} \times 3$,
where $T_c$ is the number of frames in a clip, $N_{lm} = 75$ is the total number of landmarks extracted per frame, and 3 is the number of spatial dimensions. While the x and y dimensions provide accurate positional information, the z dimension, representing depth, is prone to inaccuracies.


\subsection{Generation of the Synthetic Signer}\label{sec:synt}
The 3D ground truth from 3D-LEX is used to create synthetic data to expand the NGT200 dataset. We retarget 3D-LEX animation files onto an avatar (A) generated with \textit{Ready Player Me Studio} and rendered using the open-source framework \textit{Babylon.js}. Each animation clip is recorded from the screen in the browser from three perspectives matching the signCollect system's camera angles. We then extract landmarks from the synthetic signer using the Holistic MediaPipe framework, as described in Section \ref{Sec:stpc}.  Further details on the process for producing the synthetic videos are provided in Appendix \ref{sec:appsynth}.

\subsection{Landmark Detection Validation}\label{sec:quality}
Landmark detection can fail due to occlusions or rapid movements that blur body parts. We assess pose estimation quality by calculating the average ratio of successful to unsuccessful landmark extractions per sign. When extraction fails, MediaPipe returns a zero value for the affected keypoint. The NGT200 dataset has an average success rate of \texttt{97.6\%} for real videos and \texttt{97.8\%} for avatar videos. Detailed ratios are shown in Figure \ref{fig:kp_est} in the Appendix \ref{Sec:PQVS}.


\subsection{Dataset Characteristics}\label{Sec:characteristics}
Three native NGT signers and one synthetic signer contributed to the NGT200 dataset. Two signers consented to release videos and poses, while the third consented only to poses. Each signer was assigned a unique identifier, as detailed in Table \ref{tab:identifiers}. 
\begin{table}[!h]
\caption{Details on the availability of NGT200 modalities.}
\label{tab:identifiers}
\vskip 0.15in
\begin{center}
\begin{small}
\begin{sc}
\begin{tabular}{lllll||l}
\toprule
\textbf{ Signer ID} & 1 & 2 & 3 & A & Total\\ \toprule
\# Videos & 600 & $\times$ &600&600 & 1,800 \\
\# Poses & 600 &600&600&600 & 2,400 \\
\bottomrule
\end{tabular}
\end{sc}
\end{small}
\end{center}
\vskip -0.1in
\end{table}

Understanding the shapes and linguistic structures within data distributions can improve the design of inductive biases for ISR methods \cite{Kezar_2023, ranum2024}.
We provide some linguistic information to inform on the data distribution in the NGT200 dataset. Table \ref{tab:handedness} details the handedness distribution from SignBank: Class \textit{1} includes one-handed signs, Class \textit{2a} includes asymmetrical two-handed signs (non-dominant hand as location), and Class \textit{2s} includes symmetrical two-handed signs (both hands moving with the same handshape) \cite{Crasborn2020}. Notably, signers may not consistently use their dominant hand for strong handshapes, suggesting flip symmetries as a potential data augmentation technique. However, caution is needed as some signs encode directionality, and we have not yet assessed sign directionalities in the NGT200 dataset.
 


\begin{table}[!h]
\caption{Handedness of signs in the NGT200 vocabulary. There are in total 122 one-handed signs and 78 two-handed signs.}
\label{tab:handedness}
\vskip 0.15in
\begin{center}
\begin{small}
\begin{sc}
\begin{tabular}{lccc}
\toprule
\textbf{Handedness} & \textit{1} & \textit{2s} & \textit{2a} \\
\midrule
\textbf{Count} &122&63&15 \\
\bottomrule
\end{tabular}
\end{sc}
\end{small}
\end{center}
\vskip -0.1in
\end{table}

\begin{figure}[h!]
\vskip 0.2in
\begin{center}
\centerline{\includegraphics[width=\columnwidth]{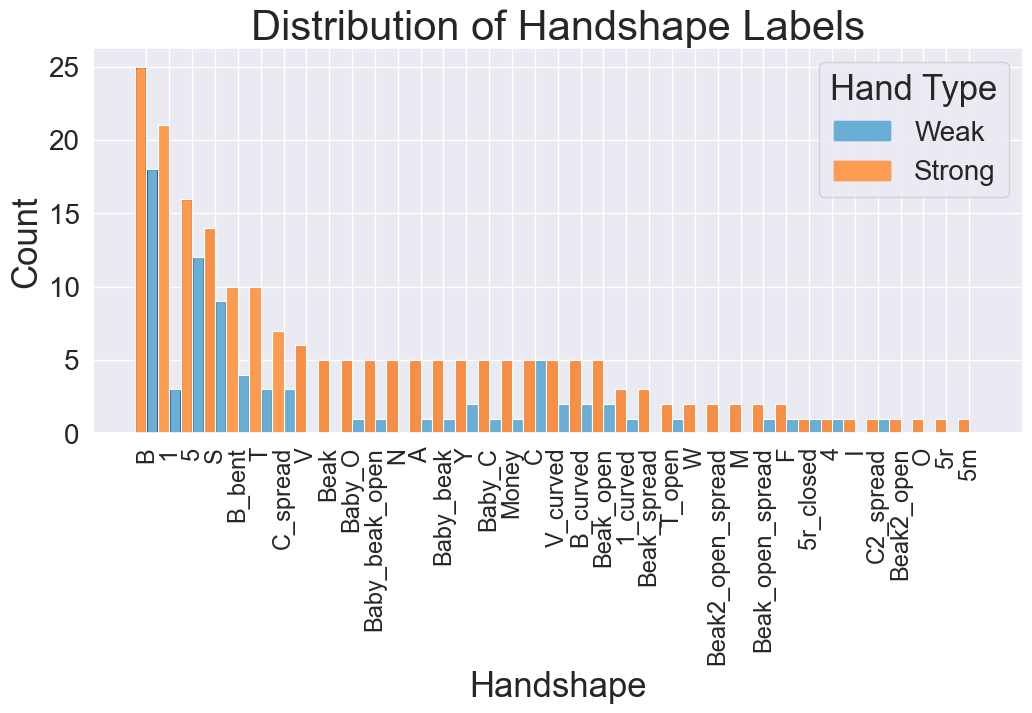}}
\caption{The frequency of each handshape type within the NGT200 vocabulary, categorized by strong (dominant hand) and weak (non-dominant hand).}
\label{fig:disthandshapes}
\end{center}
\vskip -0.2in
\end{figure}

The distribution of handshapes in the NGT200 vocabulary is illustrated in Figure \ref{fig:disthandshapes}. 
Each sign in the NGT200 vocabulary is annotated with a specific handshape for the dominant hand. Additionally, there are 78 two-handed signs, where labels are provided for the non-dominant hand as well. 

\section{The MV-ISR Task}\label{Sec:mvisr}
We now shift our focus toward the second contribution of this work: we provide a proof-of-concept demonstrating that the MV-ISR task is distinct from that of SV-ISR. We conducted a series of experiments using the NGT200 dataset which we learned from and tested on different sets of views. Specifically, we addressed the following questions:
\begin{itemize}[noitemsep] 
    \item[\textbf   {Q1:}] 
    \textit{Does viewing angle matter in pose-based ISR?}
    \item[\textbf{Q2:}] \textit{How does the inclusion of additional views during training impact performance?}
\end{itemize}
To address these questions, we use a state-of-the-art Sign Language Graph Convolution Network (SL-GCN) \cite{Jiang2021}. We construct pose graphs from the point clouds and train the SL-GCN to predict glosses.

\subsection{Method \& Experiments}

\begin{figure}[!h]
\begin{center}
\centerline{\includegraphics[width=0.85\columnwidth]{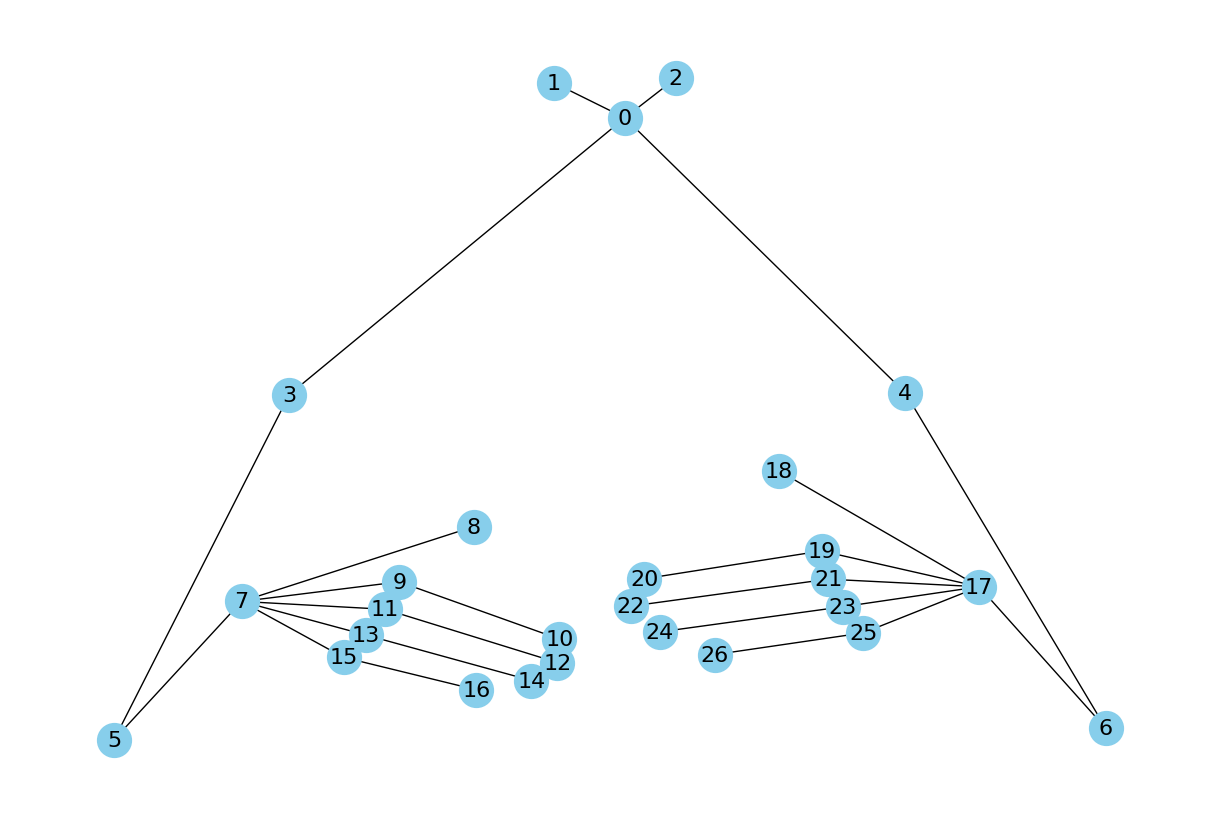}}
\caption{The reduced spatial graph used in our experiments. The graph reflects a simplified human skeleton using 27 nodes: 10 nodes per hand, and 7 nodes for the overall pose position. Spatial edges connect nodes to approximate the human bone structure. 
}
\label{fig:reduced_graph}
\end{center}
\vskip -0.2in
\end{figure}

\paragraph{Graph Construction}\label{Sec:Graph}
We adopt the graph reduction scheme introduced by \citet{Jiang2021} to mitigate noise from the numerous nodes and edges in a human skeleton and to reduce distances within the graph. We downsample to 27 nodes: 10 per hand and 7 for the overall pose. We configure the spatial edges to approximate the human bone structure, as illustrated in Figure \ref{fig:reduced_graph}.  A spatial pose graph is constructed for each frame, and the ordered graph sequence represents one sign. In this work, we do not explicitly include temporal edges, which connect the same node between consecutive frames. Instead, we manage the temporal dynamics through 1D convolutions over the time axis.

\paragraph{The SL-GCN}\label{Sec:GCN}
SL-GCN is a state-of-the-art model for pose-based ISR, featuring 
a sophisticated design engineered towards the SLR task. It builds on a spatio-temporal GCN with a spatial partition strategy to model dynamic skeletons. Furthermore, the network is enhanced with decoupled spatial convolution layers, a spatial, temporal, and channel-wise attention module, a temporal convolutional layer, and a DropGraph module. In total, 10 spatio-temporal GCN blocks are used, followed by global average spatio-temporal pooling before classification with a fully connected layer. We use the Openhands \cite{2021_openhands} implementation, featuring an SL-GCN encoder with a fully-connected classification decoder.

\begin{figure}[!h]
\vskip 0.2in
\begin{center}
\centerline{\includegraphics[width=.8\columnwidth]{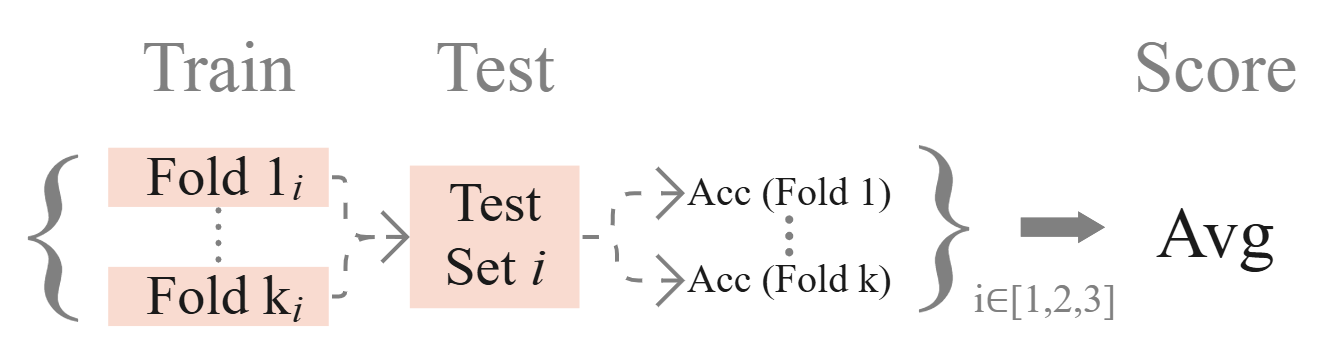}}
\caption{Performance evaluation scheme using k-fold cross-validation across three distinct test sets. Accuracy scores (Acc) are computed for each fold within a test set, and the average accuracy (Avg) is calculated across all folds.}
\label{fig:splits}
\end{center}
\vskip -0.2in
\end{figure}

\paragraph{Training Details}\label{sec:details}
We construct three train-validation-test split-blocks, configured according to the k-fold cross-validation scheme illustrated in Figure \ref{fig:splits}.  Each block has a distinct test set and train-validation splits. Each test set includes a novel human signer for a given sign, but the signer appears in the training set for other signs. Synthetic data and the SignBank video are excluded from test-set. The k-value is adjusted based on available data, detailed in Table \ref{tab:splits}. Single-view models use a three-fold cross-validation scheme, while models incorporating two or three views use a six-fold cross-validation scheme.  The final score is the average across all eighteen folds.

\begin{table}[!h]
\caption{Details on the allocation of train-validation-test examples per number of included training views. If the SB front view is included, the number of training examples increases with one. }
\label{tab:splits}
\vskip 0.15in
\begin{center}
\begin{small}
\begin{sc}
\begin{tabular}{llll}
\toprule
 \# Views&\# Train & \# Val & \# Test \\
\midrule
1 &2 (+1)$_{Sb}$&1&1 \\
2  &5 (+1)$_{Sb}$&1& 1\\
3  &7 (+1)$_{Sb}$&2&1 \\
\bottomrule
\end{tabular}
\end{sc}
\end{small}
\end{center}
\vskip -0.1in
\end{table}



\paragraph{Experiments}
We first examine whether models trained exclusively on single views maintain accuracy when evaluated on novel views. Next, we retrain a front-view model with the SignBank front-view data (ID: S$_b$)  to assess how the additional front-view data impacts performance. 
To assess the impact of multi-view data on performance, we train models on individual views and then progressively integrate additional views.

\subsection{Results}\label{Sec:mv_results}
Table \ref{tab:mvexp_f} presents the results from testing models trained with one view on all available views. When SV models are evaluated on novel views, the accuracy exhibits a relative drop of more than \texttt{50\%}. Including the extra $S_b$-view leads to a relative increase of \texttt{122\%} on the front-view, but only marginally improves side-view predictions. These findings suggest that more single-view data alone is not the solution for the MV-SLR task, as it doesn't help the model generalize across different viewpoints. 
In conclusion, the results indicate that viewing angle matters in MV-SLR and establishes MV-ISR as a distinct task from SV-ISR.

Table \ref{tab:mvexp} presents the results from incorporating additional views into the training dataset. Adding more views consistently enhances recognition accuracy across all perspectives, suggesting that the network learns distinct and complimentary features from each view.



\begin{table}[!h]
\vskip 0.15in
\begin{center}
\begin{sc}
\begin{tabular}{llllccc}
\toprule
 Train & Test  & Top1 & Top3   \\
View &  View  &  Acc &  Acc  \\
\midrule
l$^{123A}$ & l$^{123}$ & \textbf{.05}$_{(\pm.03)}$ & \textbf{.10}$_{(\pm.04)}$  \\
& f & .03$_{(\pm.01)}$ & .07$_{(\pm.03)}$ \\
 & r & .01$_{(\pm.01)}$ & .04$_{(\pm.01)}$  \\ \midrule
 r$^{123A}$ & l$^{123}$ & .02$_{(\pm.01)}$ & .05$_{(\pm.02)}$  \\
& f &  .03$_{(\pm.01)}$ & .09$_{(\pm.03)}$  \\
 & r & \textbf{.06}$_{(\pm.02)}$ & \textbf{.12}$_{(\pm.04)}$  \\ \midrule
f$^{123A}$ & l$^{123}$ & .03$_{(\pm.02)}$ & .07$_{(\pm.02)}$  \\
& f &  \textbf{.09}$_{(\pm.02)}$ & \textbf{.20}$_{(\pm.03)}$ \\
 & r & .03$_{(\pm.01)}$ & .10$_{(\pm.02)}$\\ \midrule
 f$^{123AS_b}$ & l$^{123}$ & .06$_{(\pm.02)}$ & .14$_{(\pm.04)}$  \\
& f & \textbf{.20}$_{(\pm.03})$ & \textbf{.36}$_{(\pm.05)}$  \\
 & r & .05$_{(\pm.02)}$ &  .13$_{(\pm.03)}$  \\ 
 \bottomrule
\end{tabular}
\end{sc}
\caption{Classification accuracy with standard deviation for the SL-GCN model, trained on a single view and tested across all views (L: left; R: right; F: front). The superscripts $I \in {1,2,3,A, Sb}$ indicate the signer identities associated with the training or test views. If no view or superscript is specified, the value remains the same as in the row above. Results highlighted in bold denote the top-performing view for each model.}
\label{tab:mvexp_f}
\end{center}
\vskip -0.1in
\end{table}

\begin{table}[!h]
\vskip 0.15in
\begin{center}
\begin{sc}
\begin{tabular}{llllccc}
\toprule
 Train & Test  & Top1 & Top3   \\
Views &  View  &  Acc &  Acc  \\
\midrule
lf$^{123A}$ & l$^{123}$  &  .25$_{(\pm.05)}$  &  .51$_{(\pm.06)}$  \\
lr &   & .27$_{(\pm.05)}$  & .47$_{(\pm.07)}$   \\
\textbf{lfr} &   &  \textbf{.46}$_{(\pm.04)}$  &  \textbf{.69}$_{(\pm.03)}$    \\  \midrule
lf$^{123A}$   & f$^{123}$&  .35$_{(\pm.05)}$  &  .59$_{(\pm.06)}$  \\
fr &   & .42$_{(\pm.05)}$  &  .67$_{(\pm.05)}$   \\
\textbf{lfr} &   & \textbf{.49}$_{(\pm.03)}$  &  \textbf{.74}$_{(\pm.03)}$   \\  \midrule
lr$^{123A}$ &  r$^{123}$ & .28$_{(\pm.05)}$  & .51$_{(\pm.06)}$  \\
fr &   &.39$_{(\pm.05)}$  &.62$_{(\pm.06)}$   \\
\textbf{lfr} & & \textbf{.47}$_{(\pm.04)}$  &  \textbf{.72}$_{(\pm.03)}$   \\ 
\bottomrule
\end{tabular}
\end{sc}
\caption{Classification accuracy with standard deviation from the SL-GCN using combinations of views. The highlighted results indicate the top-performing model per test-view.
\label{tab:mvexp}}
\end{center}
\vskip -0.1in
\end{table}


\section{Scaling Up Sign Language Datasets with Synthetic Data for ISR}\label{sec:synthdatasec}
In the preceding section, we utilized pose data from both human and synthetic signers. However, the impact of leveraging synthetic data to support SLP tasks is uncertain. We ask the following question:

\vspace{-1mm}
\begin{itemize}[noitemsep] 
    \item[\textbf{Q3:}] \textit{Can synthetic data be effectively used to supplement MV-SL datasets in the context of boosting pose-based MV-ISR performance?}
\end{itemize}
To address this question, we conduct experiments by iteratively augmenting the training dataset with synthetic and human signers to assess the impact of including synthetic pose data.

\subsection{Method \& Experiments}
In the experiments in this section, we reuse the graph construction method and the SL-GCN described in Section \ref{Sec:Graph}. 

\paragraph{Training Details} To provide an additional perspective on the MV-ISR task, we redefine the test set in this section to use only signer 3 for testing. Training is then conducted using exclusively poses from signers 1, 2, A, and Sb. This train-validation-test split allows us to evaluate the model's performance in the context of a novel signer prediction task, which is considered significantly more challenging than predicting signs from signers that has been seen during training.

\paragraph{Experiments} 
We iteratively augment the training data subsets to include additional signer identities and views. The experiments evaluate if: 
\begin{enumerate}
    \item[i] The addition of a synthetic signer improves overall gloss recognition accuracy
    \item[ii] There is a performance difference between adding synthetic poses and human poses
\end{enumerate}
\subsection{Results}
Table \ref{tab:mvexp_synthetic} presents the results of experiments on including synthetic data to boost the recognition performance of an MV-ISR model. The results show that adding a single frontal view from either the avatar or SignBank data improves performance across all three views. The difference between using the human signer from SignBank and the synthetic data is marginal, with human signer data providing a slightly higher improvement. Additionally, incorporating more views from the synthetic signer significantly increases recognition accuracy from the baseline, with the best results achieved by leveraging all available data.

\begin{table}[!h]
\vskip 0.15in
\begin{center}
\begin{sc}
\begin{tabular}{llll}
\toprule
Train&Test  & Top1 & Top3  \\
Views&  View  &  Acc &  Acc \\
\midrule
lfr$^{12}$&l$^{3}$& .03$_{(\pm.01)}$  & .08$_{(\pm.02)}$\\ 
&\small{f}& .14$_{(\pm.04)}$  & .27$_{(\pm.04)}$         \\
&\small{r}& .14$_{(\pm.02)}$ & .27$_{(\pm.04)}$ 
   \\ \noalign{\smallskip} \hline \noalign{\smallskip} 
lfr$^{12}$+f$^{A}$&\small{r$^{3}$}& .09$_{(\pm.02)}$   & .17$_{(\pm.04)}$  \\
&\small{f}&  .27$_{(\pm.03)}$   &.43$_{(\pm.04)}$ \\
&\small{l}& .22$_{(\pm.02)}$   & .39$_{(\pm.03)}$   \\ \noalign{\smallskip} \hline \noalign{\smallskip} 
lfr$^{12}$ + f$^{Sb}$&\small{l$^{3}$}&   .10$_{(\pm.02)}$   & .20$_{(\pm.04)}$   \\
&\small{f}& .28$_{(\pm.04)}$   & .45$_{(\pm.04)}$  \\
&\small{r} &  .26$_{(\pm.03)}$   & .41$_{(\pm.04)}$  \\ \noalign{\smallskip} \hline \noalign{\smallskip}
lfr$^{12A}$&\small{l$^{3}$}&  .19$_{(\pm.02)}$   &  .34$_{(\pm.03)}$    \\
 &\small{f}& .43$_{(\pm.02)}$   & .61$_{(\pm.02)}$    \\
&\small{r}& .38$_{(\pm.04)}$  & .57$_{(\pm.03)}$     
\\ \noalign{\smallskip} \hline \noalign{\smallskip} 
lfr$^{12A}$ + f$^{Sb} $&\small{l$^{3}$}&  .32$_{(\pm.04)}$   & .49$_{(\pm.04)}$  \\
 &\small{f}& .48$_{(\pm.02)}$   & .68$_{(\pm.02)}$     \\
&\small{r}& .43$_{(\pm.02)}$  & .60$_{(\pm.02)}$    \\
\bottomrule
\end{tabular}
\end{sc}
\caption{Classification accuracies for MV-ISR SL-GCN experiments with and without synthetic data, tested across different views. Accuracies are averaged over 10 runs with standard deviations. Experiments are trained on all 3 views from signers 1 and 2, and where indicated, 1 frontal view from SignBank and 1-3 views of the synthetic avatar. All models are tested on signer ID 3. \label{tab:mvexp_synthetic}
}
\end{center}
\vskip -0.1in
\end{table}

Furthermore, the experiments using the LFR$^{12A}$ train set in Table \ref{tab:mvexp_synthetic} are equivalent in size to the LFR$^{123A}$ experiments in Table \ref{tab:mvexp}. There is a drop of \texttt{27\%}, \texttt{6\%} and \texttt{9\%} in accuracy for the left, front, and right views, respectively, demonstrating that predicting the signs of a novel signer is indeed a more challenging task.

These findings provide empirical evidence that synthetic data can substantially inform recognition models when training with a pose modality. This is an important observation for the SLR community, suggesting a viable approach to scale up multi-view pose-based datasets to make practical applications of recognition models more feasible. We conclude that adding synthetic data boosts sign recognition accuracy in the pose modality and that including synthetic data in the NGT200 dataset is a beneficial strategy for enhancing overall model performance.

\section{The Case for Geometric MV-SLR}\label{Sec:GeomSLR}
As observed in Section \ref{Sec:mv_results}, the SL-GCN achieves a top recognition accuracy of \texttt{49\%} in our experiments on the NGT200 dataset. To explore more efficient learning in the context of MV-ISR, we propose leveraging geometrically grounded models. To take a first step in this direction, we ask the question:
\begin{itemize}[noitemsep] 
    \item[\textbf{Q4:}]\textit{Is a geometrically grounded model viable for ISR?}
\end{itemize}
To address this question, we modify a SE(2)-equivariant neural network proposed by \citet{bekkers2024fast}. We explore the possibility of leveraging equivariance towards the group of roto-translations in the 2D plane to enhance learning across intra-view inter-signer variations. Exploration of models equivariant towards the group of perspective transformations, and more appropriately addressing inter-view variations, is left for future work.

\subsection{Method \& Experiments}
\paragraph{The PONITA Architecture}\label{sec:ponita}
PONITA is a general purpose SE(N)-Equivariant model proposed by \citet{bekkers2024fast}, which achieves state-of-the-art results in tasks including interatomic potential energy prediction, trajectory forecasting in N-body systems, and molecule generation. They formalize the notion of weight sharing in convolutional networks as the sharing of message functions over point-pairs that should be treated equally. They derive practical pair-wise attributes that uniquely identify such equivalence classes of point pairs, and subsequently use them to build efficient equivariant architectures. 

\begin{figure}[ht]
\vskip 0.2in
\begin{center}
\centerline{\includegraphics[width=0.65\columnwidth]{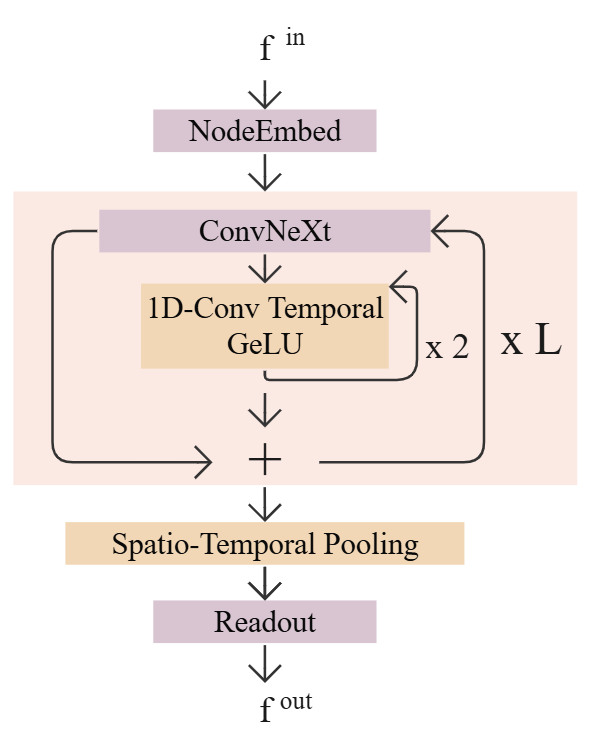}}
\caption{The modified PONITA architecture with temporal learning mechanisms (temporal-PONITA) first embeds the input features with a linear layer, which are then passed through L temporal-PONITA layers. Each layer includes one ConvNeXt block and one temporal block comprising two convolutional layers with GeLU activations.}
\label{fig:ponita}
\end{center}
\vskip -0.2in
\end{figure}

To adapt PONITA to the SLR task, we make two modifications to the model architecture. The PONITA architecture including our modifications (temporal-PONITA) is summarized in Figure \ref{fig:ponita}.

\begin{enumerate}[]
    \item[i] After each spatial PONITA layer we add a temporal convolution module consisting of two convolution kernels, GeLU activations and a residual connection.
    \item[ii] After the final temporal convolution block in the last layer, we add a spatio-temporal pooling across each pose-graph. 
\end{enumerate}

\paragraph{Experiments}
To establish a benchmark for geometric models in the MV-ISR task, we reproduce a subsection of the experiments conducted in Section \ref{Sec:mvisr}, replacing the SL-GCN with our temporal-PONITA.  
Hyperparameters and training conditions are detailed in Appendix \ref{sec:detparams}.

\begin{table}[!h]
\vskip 0.15in
\begin{center}
\begin{sc}
\begin{tabular}{llllccc}
\toprule
 Train & Test  & Top1 & Absolute   \\
Views &  View  &  Acc &   Gain \\
\midrule
lf$^{123A}$ & l$^{123}$ &   .43$_{\pm.03}$  & \textcolor{darkgreen}{+.18}  \\
lr &  &  .48$_{\pm.03}$  & \textcolor{darkgreen}{+.21}   \\
lfr &  & \textbf{.54}$_{\pm.03}$  &  \textcolor{darkgreen}{+.08} \\  \midrule
lf$^{123A}$ & f$^{123}$ &   .55$_{\pm.03}$  &  \textcolor{darkgreen}{+.20} \\
fr &  & .57$_{\pm.02}$  &  \textcolor{darkgreen}{+.15}  \\
lfr &  &  \textbf{.59}$_{\pm.03}$  &  \textcolor{darkgreen}{+.10}   \\  \midrule
lr$^{123A}$ & r$^{123}$ &  .50$_{\pm.03}$  & \textcolor{darkgreen}{+.22}  \\
fr &  &  .49$_{\pm.02}$  &  \textcolor{darkgreen}{+.10} \\
lfr &  & \textbf{.55}$_{\pm.02}$  &  \textcolor{darkgreen}{+.08}\\
\bottomrule
\end{tabular}
\end{sc}
\caption{ Classification accuracies and model standard deviations achieved with temporal-PONITA using different combinations of views. The \textit{Absolute Gain} column indicates the absolute improvement of the results presented here compared to the Top-1 accuracies in Table \ref{tab:mvexp}.
\label{tab:ponita_results}}
\end{center}
\vskip -0.1in
\end{table}

\subsection{Results}
Table \ref{tab:ponita_results} showcases the recognition accuracy of temporal-PONITA when trained and tested on NGT200. The experiments in this table correspond to those in Table \ref{tab:mvexp}, and the \textit{Absolute Gain} refers to the improvement in Top-1 accuracy compared to Table \ref{tab:mvexp}. Temporal-PONITA achieves higher performance across all combinations of views compared to SL-GCN. Additionally, temporal-PONITA demonstrates higher efficiency in terms of speed and exhibits more stable training runs with lower variance in predictions. An example comparing the training of temporal-PONITA and the SL-GCN is available in the Appendix Section \ref{Sec:AC}. These results confirm that geometrically grounded models are viable for training MV-ISR models, offering considerable benefits. 



\section{Discussion and Conclusion}

NGT200 is a pose-based dataset, which is less common than standardized video datasets. The pose modality offers advantages, such as a lower-dimensional representation of signs that generalizes better to unseen signers and backgrounds. It can embed a skeletal inductive bias into SLP models by constructing spatial edges that mirror natural human body connections \citep{Saunders2021}. Ethically, it enhances signer anonymity compared to video. However, the pose modality has limitations in SLR due to the landmark estimation process, leading to information loss, especially when considering interacting body parts \citep{moryossef2021}.


One of the main limitations of the NGT200 dataset is its size and scope. While NGT200 serves as a valuable research dataset, it is not suitable for training real-world ISR systems. In future work, we aim to expand NGT200 to include new signers, larger vocabularies, and continuous signing. The NGT200 dataset has some technical limitations: a few videos are missing due to issues occurring during collection. Additionally, the synthetic data was recorded in the browser, which occasionally can result in a lower-than-normal frame rate. Details are provided in Appendix \ref{Sec:PQVS}.

We evaluated the use of synthetic data by including poses estimated from a sign avatar. Our promising results indicate the potential of synthetic data in SLP, but its effectiveness in the RGB modality, which poses specific challenges related to signer appearance, remains unclear. Furthermore, we have not evaluated whether synthetic data might hinder the model's ability to learn authentic SL features at scale. Synthetic data could introduce false motion patterns and unrealistic articulation styles, potentially affecting real-world recognition performance. With the availability of larger multi-view datasets, the impact of synthetic data should be reassessed, especially for methods involving larger data distributions and continuous signing. Furthermore, capturing 3D ground truths can be costly. Future research should consider using 3D ground truths to generate variations in synthetic data, such as different sign amplitudes and articulation styles. Our study highlights the early stage of synthetic SL representations and raises the question of how to optimally leverage synthetic data to support SLP tasks.

We took a first step in assessing the potential of GDL tools for supporting SLP tasks by demonstrating the application of an SE(2)-equivariant model, which achieved significant improvements over the baseline. However, many geometrically grounded methods may be better suited for SLP tasks. Future work should consider, \textit{e.g.}, models equivariant to 2D perspective transformations for MV-ISR/SLR.

In this work, we highlighted the importance of viewing angles in MV-ISR. Our contributions include: i) the NGT200 dataset; ii) demonstrating that recognition models trained on frontal views lose accuracy on side views, and showing that MV recognition accuracy improves when learning from multiple views; iv) enhancing the NGT200 dataset with synthetic multi-view data, which demonstrates the potential for scaling up multi-view datasets; and v) showcasing the benefits of considering geometrically grounded models for MV-ISR tasks. We hope this dataset will benefit the research community by providing a foundation for exploring a novel and intriguing task in geometric deep learning, inspiring new and stronger approaches to SLP.

\section{Privacy and Ethical Considerations}\label{sec:ethics}
The increasing demand for data to drive computational methods and machine learning algorithms introduces significant privacy risks and ethical concerns across the computational sciences. These issues are particularly pronounced in data collection involving minority groups, such as sign language communities. \citet{Bragg2020} emphasizes that gathering data from small populations inherently reduces anonymity. Another critical issue is the collection of data without obtaining informed consent from contributors. In the case of the NGT200 dataset, all participants provided informed consent and received compensation. The video modality of one signer is not released to preserve the anonymity of this signer. Names of signers are not disclosed. Instead, each signer was assigned a unique signer ID as described above.

\section{Positionality Statement and Contributions}
Research into the automatic understanding and processing of sign languages requires collaboration across multiple disciplines, bringing diverse positionalities, knowledge, and expertise into the team. Consequently, we include a brief note on our research team members and their respective contributions to this project.

Ranum is a hearing sibling of a signing adult with a language learning disability, and Norwegian SL is her second language; Otterspeer is deaf and an expert NGT signer; Roelofsen is a hearing parent of a deaf child, and proficient in NGT; Andersen is hearing with basic proficiency in NGT; Wessels and Bekkers are non-signing. 

Ranum, Roelofsen, Wessels, and Bekkers have backgrounds in Artificial Intelligence, with Roelofsen additionally having a background in linguistics. The methods considered were mostly implemented by Ranum, who also primarily authored the current manuscript. Roelofsen supervised the development of this project, providing feedback on the manuscript and contributing to discussions. Wessels and Bekkers contributed to this project with their insights and expertise on the topic of geometric principles in deep learning; Otterspeer and Andersen have a background in programming, signing avatars and system engineering for sign capture, and developed the pipeline for producing the synthetic data. Additionally, Otterspeer conducted the collection of the video data. All authors edited and commented on previous versions of the manuscript. All authors read and approved the final manuscript.

\section{Data and Code}
NGT200 is available through OSF:  \href{https://osf.io/5zuyd/}{osf.io/5zuyd/}.  The code for reproducing our experiments is available on OSF or (WIP) at GitHub:  \href{https://github.com/OlineRanum/GMVISR}{github.com/OlineRanum/GMVISR}.

\bibliography{main}
\bibliographystyle{icml2024}


\appendix
\onecolumn

\section{Training Details Supplements}\label{sec:detparams}
Table \ref{tab:hyper} details the training configurations and hyperparameters used in our experiments. We utilized the default parameters for SL-GCN as provided in the OpenHands framework and adhered to the OpenHands standard for the temporal convolution kernel in both models. For the additional parameters of the temporal-PONITA architecture, we conducted a hyperparameter sweep using Optuna, an open-source framework that automates hyperparameter tuning for machine-learning models.

\begin{table}[!h]
\vskip 0.15in
\begin{center}
\begin{sc}
\begin{tabular}{lcc}
\toprule
Hyperparameters & SL-GCN  & Temporal-PONITA \\
\midrule
Epoch strategy & Early Stopping & Early Stopping \\
Warmup & -& 100 \\
Batch size & 32&32 \\
Learning rate &1e-3&5e-3 \\ \midrule
Hidden dim &64, 128, 256& 64\\
Layers &10 & 6\\
\midrule
Temporal convolution kernel size & 9 & 9 \\
Temporal convolution weight decay & - & 1e-3 \\ \midrule
Number of orientations&-& 1\\
Basis dim &-& 128\\
Degree of polynomial embedding &-& 1\\
Widening factor &-& 4\\
Layer scale &-& 0\\
\bottomrule
\end{tabular}
\end{sc}
\caption{ Hyperparameters used in training of each model. \begin{sc}"-"\end{sc} Indicates that the parameter does not apply to this model.
\label{tab:hyper}}
\end{center}
\vskip -0.1in
\end{table}

\section{Pose Quality Validation Supplements}\label{Sec:PQVS}
\begin{figure}[ht]
\vskip 0.2in
\begin{center}
\centerline{\includegraphics[width=0.75\columnwidth]{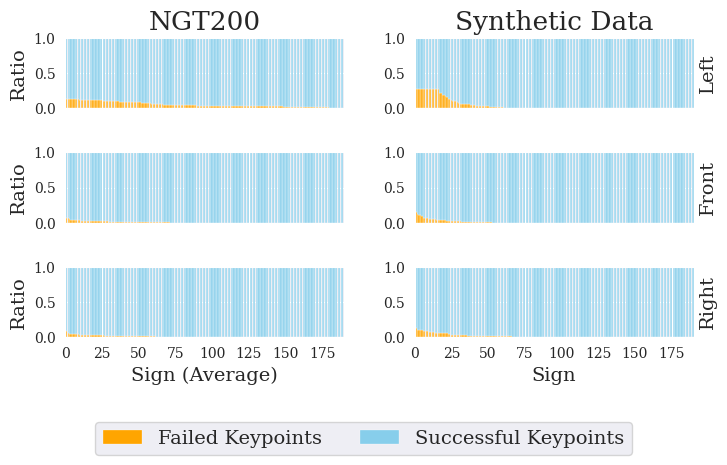}}\centerline{\includegraphics[width=0.35\columnwidth]{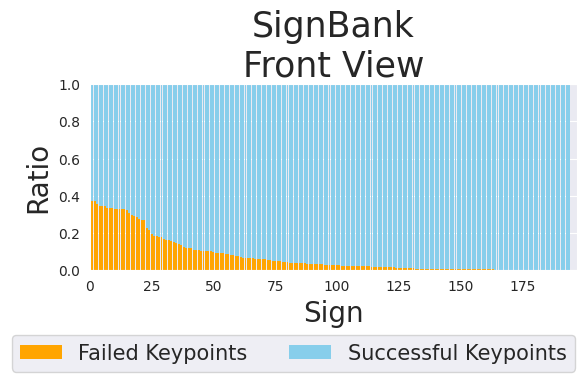}}
\caption{The average ratio of successful to failed keypoint detections for each sign across different views (Left, Front, Right) in both NGT200 and Synthetic datasets. Each bar represents the ratio for an individual sign, sorted by the magnitude of the ratio of failed to successful keypoint detections. The blue bars indicate successful keypoint detections,  whereas the orange bars represent instances where keypoints failed, with MediaPipe returning a value of zero. }
\label{fig:kp_est}
\end{center}
\vskip -0.2in
\end{figure}

Estimating accuracy in key pose extraction is challenging. To indicate the quality of the Mediapipe pose estimation process, we present the ratio of successful to unsuccessful landmark extractions from the video in Figure \ref{fig:kp_est}. Estimating accuracy in key pose extraction is challenging. To assess the quality of the Mediapipe pose estimation process, we present the ratio of successful to unsuccessful landmark extractions from the video in Figure \ref{fig:kp_est}. In this figure, successful keypoints are shown in blue, while failed keypoint extractions are shown in orange. Each bar was calculated by summing the total number of zeros occurring in each spatial graph across all consecutive time-frames within a sign, and comparing them to the total number of non-zero occurrences. 
The rates are sorted by the magnitude of failed keypoints per sign, but may not align across different subplots (e.g., the sign characterized by the first bar in the left view may not correspond to the sign characterized by the first bar in the front view). As observed, keypoint extraction from both human and synthetic data performs well, indicating acceptable dataset quality. Sources of failed keypoints may include occlusion of hands and other body parts, as well as rapid motion across consecutive frames.

\section{Details on the Synthetic Pose Production}\label{sec:appsynth}
The 3D-LEX dataset employs three distinct motion capture systems to accurately capture handshapes, facial expressions, and full-body postures directly from human signers, ensuring high fidelity to signs. Handshapes are recorded using StretchSense gloves, and full-body poses with a Vicon motion capture rig. The dataset includes animation files combining handshapes and full-body poses. For synthetic avatar videos, we extract 200 animation files from 3D-LEX that overlap with the NGT200 vocabulary, retarget this data onto the Ready Player Me avatar, and use it to display the recorded signs.

\begin{figure}[!h]
    \centering
    \includegraphics[width=.95\linewidth]{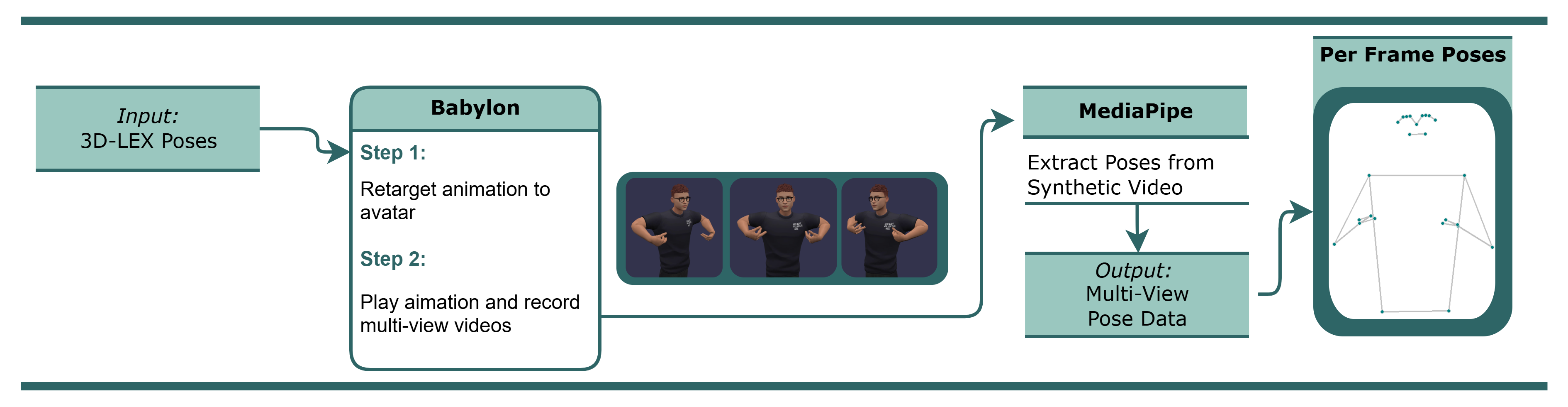}
    \caption{Pipeline for producing the synthetic animation data with Babylons.js and MediaPipe. The chart showcases the multi-view recordings captured of the Ready Player Me Avatar.}
    \label{fig:flow}
\end{figure}

To produce synthetic multi-view videos using 3D ground truth, we developed a web application using BabylonJS, an open-source web rendering engine. This application retargets motion capture data from 3D-LEX and supports batch processing of motion capture files. Within the app, the uploaded files are played sequentially, and a screen recorder captures videos of the signs from distinct viewpoints. These videos are then downloaded, and MediaPipe is used to extract poses from the synthetic videos. The pipeline, including a screenshot of the Ready Player Me avatar, is shown in Figure\ref{fig:flow}. Our (working) repository is available at  \href{https://github.com/J-Andersen-UvA/BabylonSignLab.git}{https://github.com/J-Andersen-UvA/BabylonSignLab.git}, which includes a live demo.

\section{Comparison with Other Isolated Sign Datasets and State-Of-The-Art Methods}\label{sec:comparison}

Table \ref{tab:comparison} compares the NGT200 dataset and temporal-PONITA with other commonly used isolated sign datasets and ISR State-Of-The-Art (SOTA) methods. We do not list continuous datasets, even though some sentence-level datasets do offer multiple views for advanced studies, such as \citet{Duarte_2021} and \citet{forster-etal-2014}. Note that this list is not exhaustive, but simply aims to achieve some commonly considered datasets. Among these, we are only familiar that the NGT200 and a subset of WLASL-2000 have access to 3D-ground truth. Other ASL datasets may overlap with the 3D-LEX ASL dataset, making 3D ground truth available, but this has not yet been evaluated. 

\begin{table}[!h]
    \centering 
    \caption{Overview of Datasets Used in ISR Research, and State-Of-The-Art methods. \textit{Pub. Acc.} refers to Publicly Accessible, and indicates whether the dataset is publicly available at the time of publication. In the columns \textit{Vids/Sign} we present the number of videos \textit{P/V } for multi-view datasets, meaning \textit{Videos per Sign per View}. The letter \textit{U} denotes that the information is unknown, and \textit{HH} stands for hard of hearing. If multiple views are present, the SOTA is averaged across all views. Consent is marked as unknown if the paper does not clarify whether informed consent was obtained.}
    \label{tab:comparison}
    \vskip 0.15in
    \resizebox{\textwidth}{!}{%
        \begin{small}
            \renewcommand{\arraystretch}{1.5}
            \begin{sc}
                \begin{tabular}{p{2cm}cp{0.9cm}p{1cm}p{0.9cm}p{1.2cm}lp{1.7cm}lcp{2cm}}
                    \toprule
                    \textbf{Dataset} & \textbf{Pub.}& \textbf{Source} & \textbf{Vocab.} & \textbf{Vids/} &\textbf{Signers} & \textbf{\# of} & \textbf{Collection} & \textbf{Cons.} & \textbf{SOTA} & \textbf{Model}\\
                    \textbf{Name} & \textbf{Acc.}& \textbf{Lang.} & \textbf{Size} & \textbf{Sign}  && \textbf{Views} & \textbf{Method} & \textbf{Access} & \textbf{A@1} & \\\toprule
                    NGT200 &$\checkmark$  & NGT & 200 & 12 (4 P/V)& 4 Deaf &3&  Curated / Lab&  $\checkmark$&.56& Temporal Ponita*\\
                    MVSL \cite{Gao2023} & $\times$& CSL&50&210 (70 P/V)&14 U&3& Curated /  Lab & $\checkmark$ &.95 & MVTK \cite{Gao2023}\\ 
                    WLASL-2000 \cite{li2020word}&$\checkmark$&ASL&2,000&10.5&119 U&1&  Scraped & $\times$ & .56 & TMS-Net \cite{DENG2024127194}\\
                    SEM-LEX \cite{Kezar_2023}&$\checkmark$ &ASL &3,149&21& 41 Deaf&1& Curated / Web-Cam & $\checkmark$ &.87& SL-GCN \cite{kezar2023exploringstrategiesmodelingsign} \\
                    ASL Citizen \cite{ASLCIT} &
                    $\checkmark$ & ASL & 2,731&30.5&52 Deaf / HH&1&Crowd&$\checkmark$  &.63&I3D \cite{ASLCIT}\\
                    MS-ASL-1000 \cite{MS_ASL} & $\checkmark$ & ASL & 1,000& 25 & 222 U & 1 &Scraped&$\times$&.70 &MASA \cite{10549522}\\
                    INCLUDE \cite{Sridhar}&$\checkmark$& Indian SL&263&16&7 Deaf&1&Curated / Mixed Scenes&U&.81& VGG-19 + BiLSTM \cite{Das2022}\\
                    AUTSL \cite{9210578}&$\checkmark$&Turkish SL&226&170&42 Mixed\footnotemark[1]&1&Curated/ Mixed Scenes&$\checkmark$&.97& TMS-Net \cite{DENG2024127194}\\
                    MINDS-Libras \cite{Rezende2021DevelopmentAV}&$\checkmark$&Braz- ilian SL&20&60&12 Mixed&U&Curated / Lab & 1 &.97&  MIPA-ResGCN \cite{NAZ2023109009}\\
                    \bottomrule
                \end{tabular}
            \end{sc}
        \end{small}
    }
    \vskip -0.1in
\end{table}

\footnotetext[1]{Mixed includes a varying selection of instructors, translators, code, new signers, and trained signers. }

\section{Model performance comparison} \label{Sec:AC}
Training times were recorded across 10 runs for both the SL-GCN and temporal-PONITA models trained on all three views. The average time-cost calculations are provided in Table \ref{tab:compare_speed}. Although temporal-PONITA is computationally more demanding, leading to a higher average time per epoch, it is approximately \texttt{40\%} faster in terms of total running time compared to the SL-GCN.

\begin{table}[!h]
\vskip 0.15in
\begin{center}
\begin{sc}
\begin{tabular}{lcc}
\toprule
Model & SL-GCN & Temporal-PONITA \\
\midrule
Average time per epoch & 8.0 s& 11.5 s\\
Average \# epochs before stopping & 357 & 145 \\
Total time cost & 47m & 28m\\
\bottomrule
\end{tabular}
\end{sc}
\caption{Training time details of Temporal-PONITA and the SL-GCN.
\label{tab:compare_speed}}
\end{center}
\vskip -0.1in
\end{table}

\begin{figure}[!h]
    \centering

    \vspace{0.5cm}

    \subfigure[Temporal-PONITA Training Accuracy]{\includegraphics[width=0.45\linewidth]{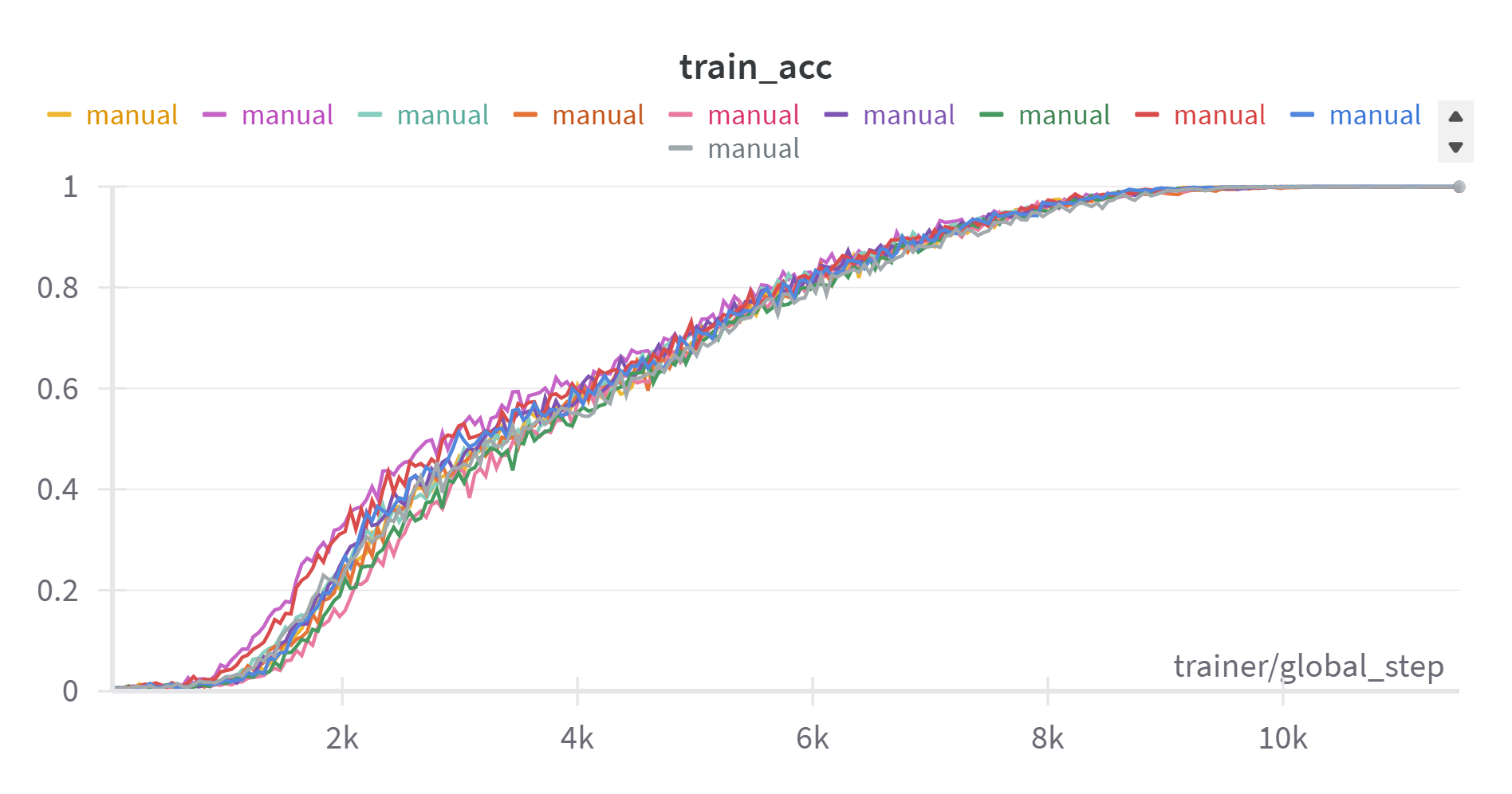}} \quad
    \subfigure[SL-GCN Training Accuracy]{\includegraphics[width=0.45\linewidth]{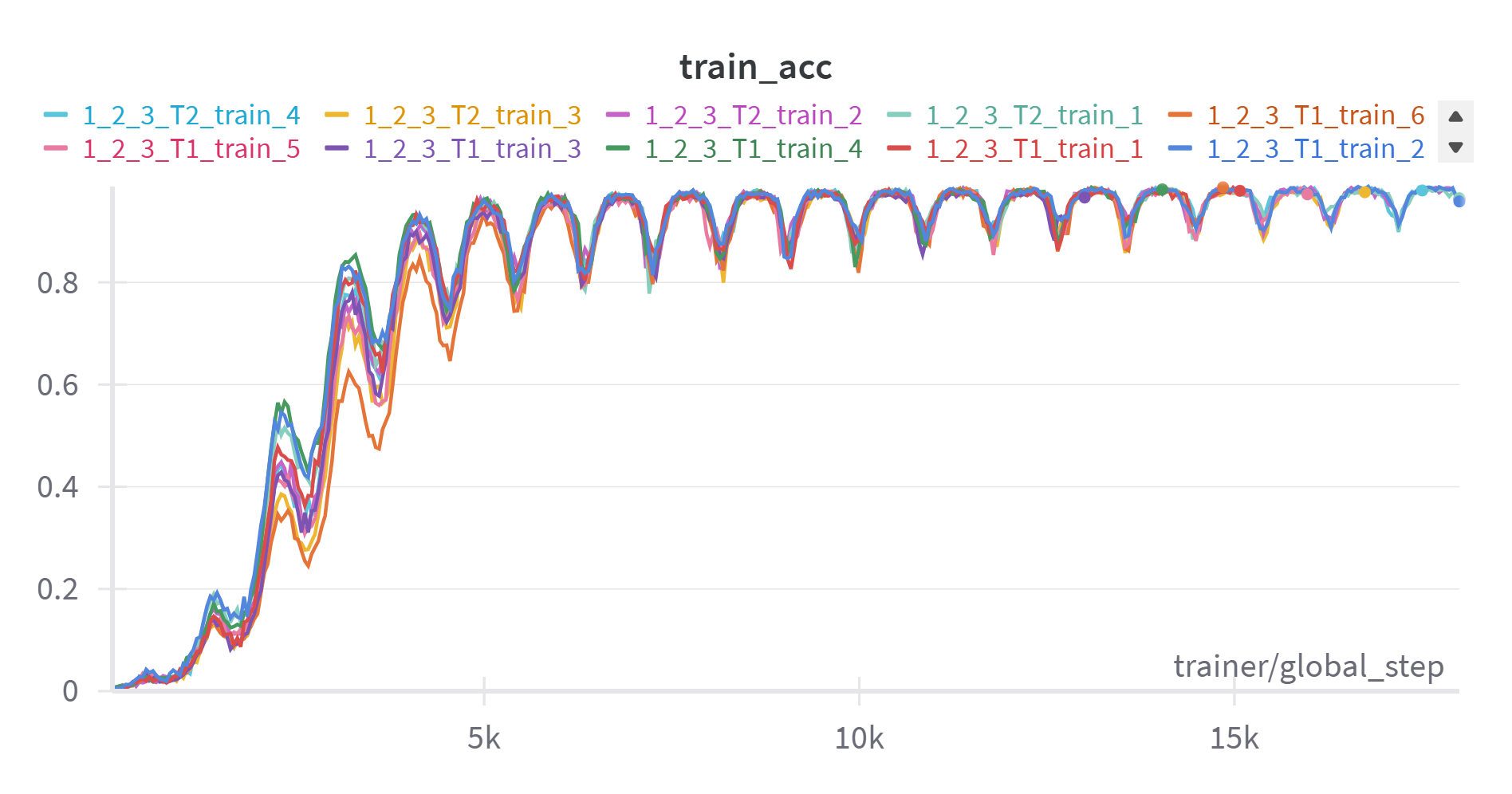}}
    
    \vspace{0.5cm}

    \subfigure[Temporal-PONITA Validation Accuracy]{\includegraphics[width=0.45\linewidth]{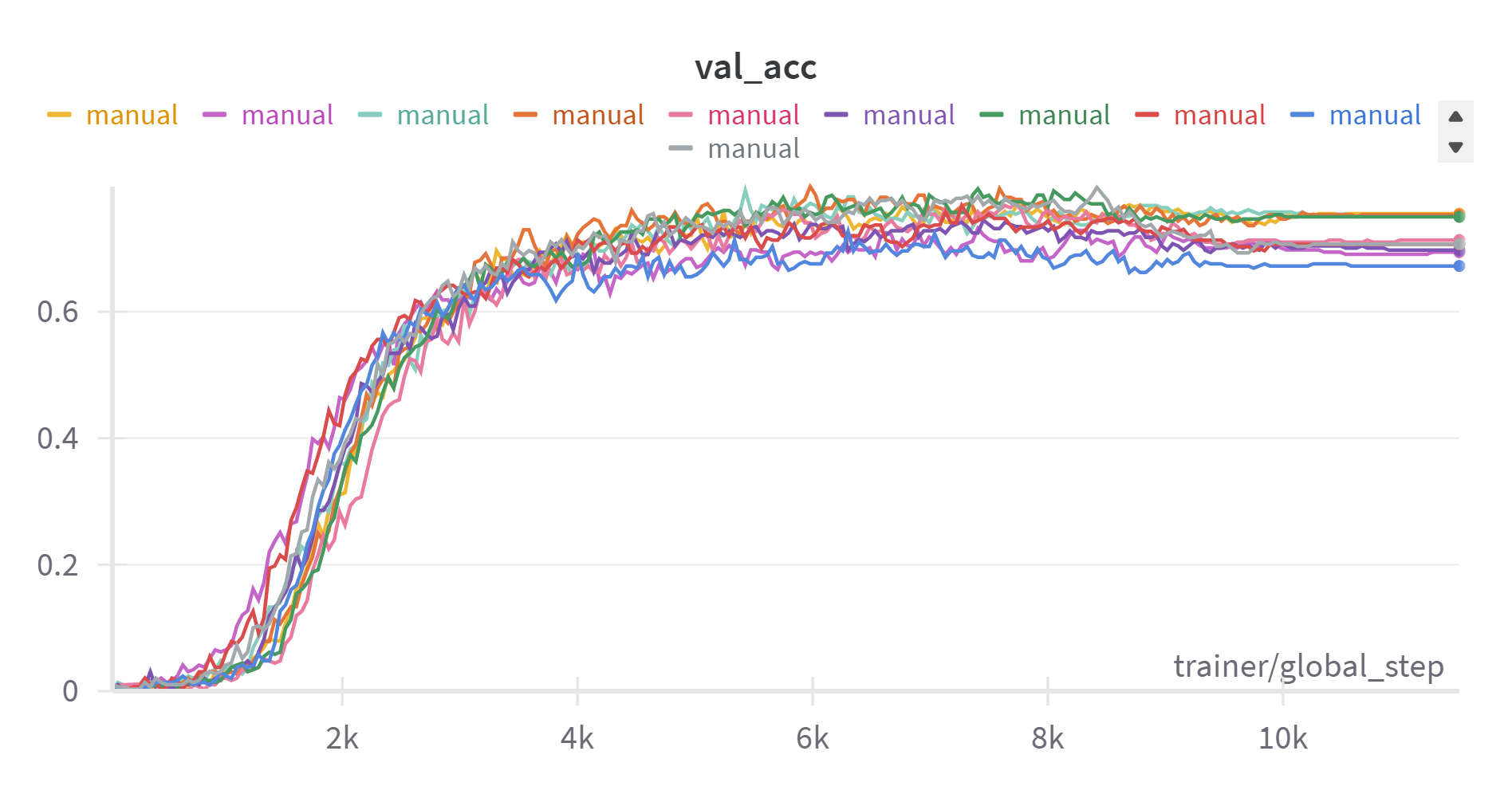}} \quad
    \subfigure[SL-GCN Validation Accuracy]{\includegraphics[width=0.45\linewidth]{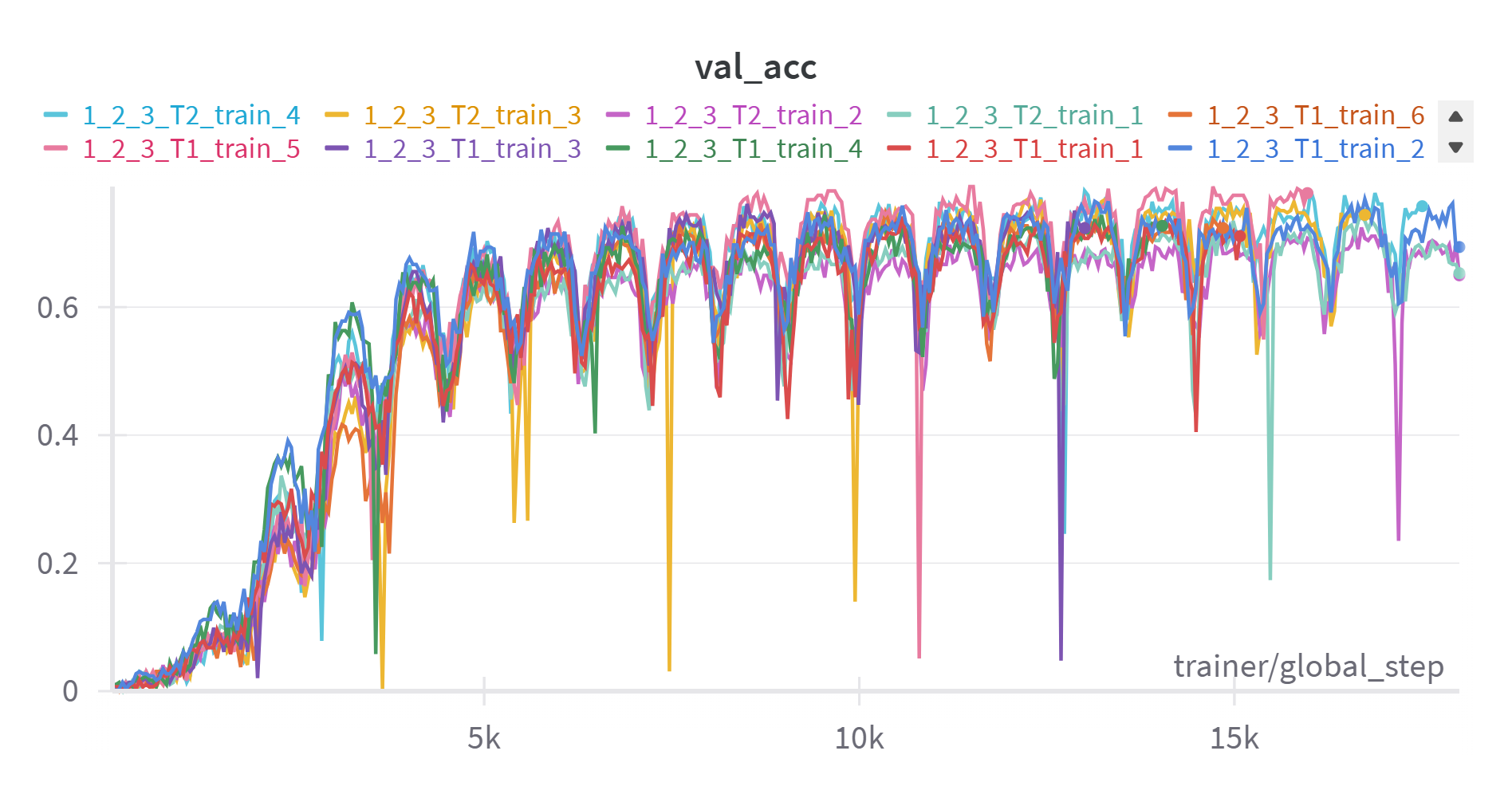}}

    \caption{Comparison of learning performance between temporal-PONITA and the SL-GCN model.}
    \label{fig:comparison}
\end{figure}

Figure \ref{fig:comparison} shows the training and validation learning curves for 10 runs in experiments involving all three views. Comparing the curves between the two models, temporal-PONITA exhibits a more stable training profile, unlike the oscillating profile of SL-GCN. Additionally, temporal-PONITA converges faster in terms of each global step to approximately the same validation accuracy as SL-GCN. However, as found in Section \ref{Sec:GeomSLR}, temporal-PONITA outperforms SL-GCN in test accuracy, indicating better generalization.

\end{document}